\documentclass{article}
\usepackage{amsmath}
\usepackage{graphicx}
\usepackage{array}
\usepackage{graphicx}
\usepackage{tabularx}
\usepackage{geometry}
\usepackage{booktabs}
\title{KunlunBaize : LLM with Multi-Scale Convolution and Multi-Token Prediction Under TransformerX Framework}

\author{Cheng Li, Jiexiong Liu, Yixuan Chen, Yanqin Jia, Zhepeng Li}
\date{}
\begin{document}

\maketitle
\begin{abstract}
Large language models have demonstrated remarkable performance across various tasks, yet they face challenges such as low computational efficiency, gradient vanishing, and difficulties in capturing complex feature interactions. To address these limitations, a novel framework has been proposed. This framework incorporates a learnable dense residual skip connection mechanism, a TransformerX module—a transformer-based component integrating multi-scale convolution and adaptive activation functions—and a multi-token prediction interaction module. The learnable dense residual connections enhance information flow and feature capture across layers. Within the TransformerX module, large convolutional kernels aggregate semantic information from extensive text segments, while smaller convolutions focus on local word order and syntactic structures. The adaptive activation function dynamically adjusts its parameters based on the semantic features of the input text, improving the model's ability to handle diverse semantic expressions and complex relationships. The multi-token prediction module boosts data utilization and accelerates inference by predicting multiple future tokens. These components significantly enhance the performance and efficiency of large language models.
\end{abstract}

\begin{figure}[htbp]
    \centering
    \includegraphics[width=\textwidth]{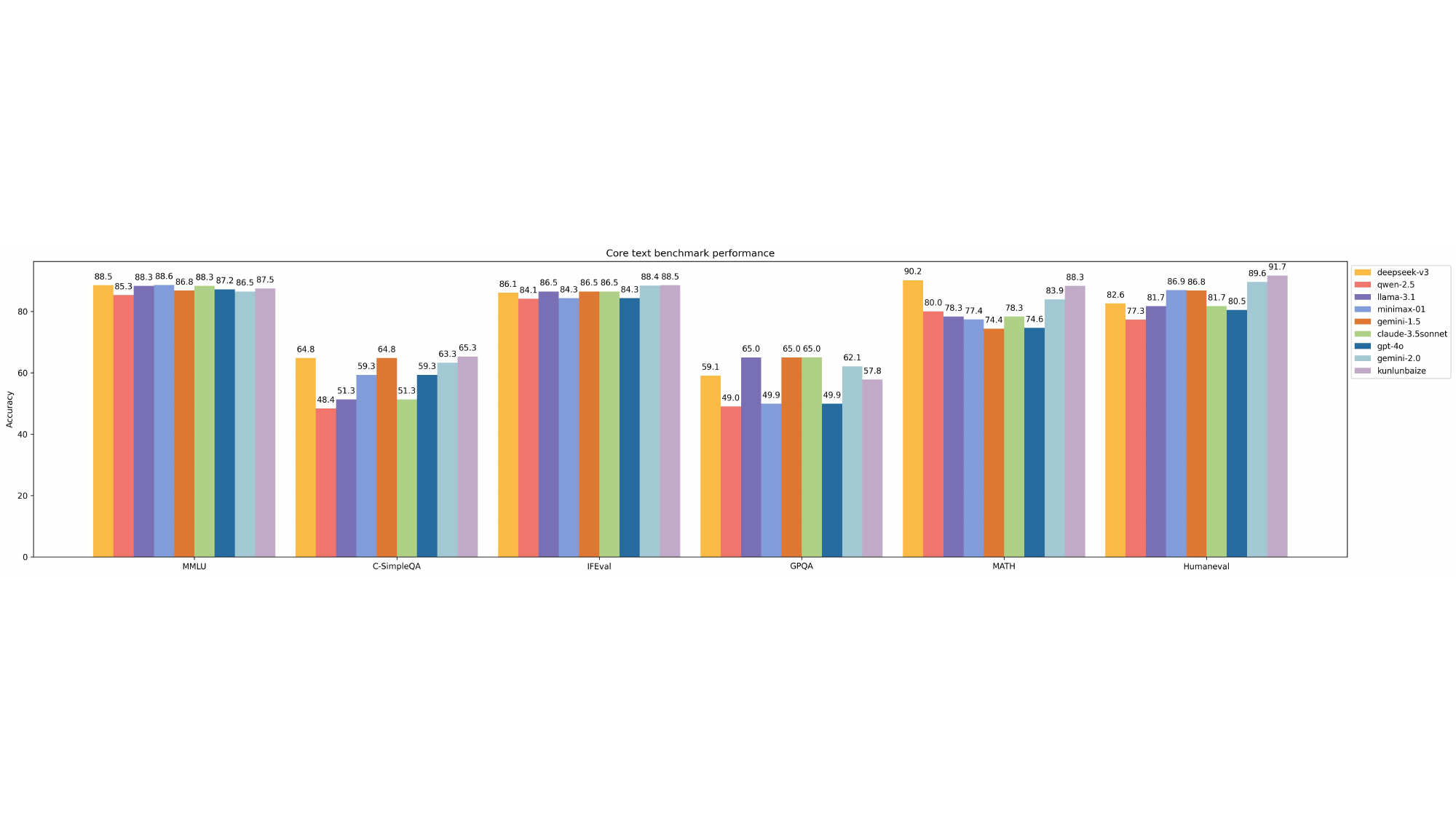}
    \caption{KunlunBaize on core text benchmarks.}
    \label{fig:your_label}
\end{figure}

\section{Introduction}

In recent years, large language models have achieved groundbreaking progress in the field of natural language processing. They have demonstrated remarkable performance across a multitude of tasks and have been widely applied in various scenarios such as intelligent customer service, content creation, and intelligent translation, profoundly transforming the way people interact with language. Evolving from early simple models based on rules and statistics to today's large-scale pre-trained models powered by deep learning, large language models have undergone several technological revolutions. As Vaswani et al. (2017)\cite{r1} proposed the Transformer architecture in "Attention Is All You Need", it has become a fundamental component in many current large language models, enabling them to handle sequential data more effectively. The GPT series of models, for instance, build upon this architecture and have shown powerful text generation capabilities, as described by Radford et al. (2018, 2019, 2020)\cite{r2,r3,r4} in their works "Improving Language Understanding by Generative Pre-training", "Language Models are Unsupervised Multitask Learners", and "Language Models are Few-Shot Learners". The BERT model, introduced by Devlin et al. (2018)\cite{r5} in "BERT: Pre-training of Deep Bidirectional Transformers for Language Understanding", has excelled in semantic understanding tasks, significantly propelling the development of applications such as question-answering systems.

However, despite their rapid development, large language models face numerous formidable challenges. Low computational efficiency has become a key factor restricting their application in resource-constrained environments. The high computational costs not only impede the rapid deployment of models but also limit their use in scenarios with high real-time requirements. As discussed by Piccialli et al. (2025)\cite{r6} in "Federated and Edge Learning for Large Language Models", the application of large language models on edge devices is hindered by their high computational demands. The problem of gradient vanishing makes it difficult for models to effectively capture long-distance dependencies during the training process. This restricts the model's ability to understand and express complex semantic information, affecting its performance when handling long texts and complex logical reasoning tasks. Additionally, as pointed out by Barredo Arrieta et al. (2020)\cite{r7} in "Explainable Artificial Intelligence (XAI): Concepts, Taxonomies, Opportunities and Challenges toward Responsible AI", understanding the decision-making processes of large language models remains a challenge due to their complex architectures. How to effectively capture the complex interactions between features at different levels and granularities remains a bottleneck in the development of large language models. For example, in practical applications, models may struggle to accurately grasp the meanings of polysemous words in different contexts or fail to fully explore the potential semantic associations within texts\cite{r23}.

To overcome these limitations and drive the further development of large language models, this paper proposes a novel framework. This framework incorporates a learnable dense residual skip connection mechanism, a TransformerX module, and a multi-token prediction interaction module. The learnable dense residual connections are designed to enhance information flow between layers and improve the model's ability to capture features at different levels, effectively alleviating the problem of gradient vanishing. The TransformerX module innovatively integrates multi-scale convolutions and adaptive activation functions into the Transformer architecture. Large convolutional kernels\cite{r8} are responsible for aggregating macro-semantic information from long text segments, while small convolutional kernels focus on local word order and syntactic structures. The adaptive activation function can dynamically adjust its parameters according to the semantic features of the input text, enabling the model to better handle diverse semantic expressions and precisely capture complex semantic relationships. The multi-token prediction interaction module significantly improves data utilization and accelerates the inference process by predicting multiple future tokens. Through the synergistic effect of these modules, we expect to significantly enhance the performance and efficiency of large language models, laying a solid foundation for their in-depth application in a broader range of fields. 

\textbf{Our main contributions are as follows:}

\begin{itemize}
    \item We propose a novel learnable dense residual skip connection mechanism. It dynamically adjusts information flow, enabling the model to capture complex inter-level feature relationships. This significantly enhances the model's expressiveness and alleviates the vanishing gradient problem, ensuring smooth training and better performance.
    \item We introduce the TransformerX module with multi-scale convolutions and the adaptive eSwish activation function. Large-scale convolutions aggregate macro-semantic information, while small-scale ones focus on local word order and syntax. The eSwish function adapts to input semantics, strengthening the model's ability to handle diverse expressions and contexts.
    \item We develop a multi-token prediction interaction module. By expanding the prediction range to multiple future tokens, it densifies the training signal and improves data utilization. It promotes feature interaction between different layers, maintaining a causal chain at each prediction depth, thus accelerating inference and enhancing overall performance.
\end{itemize} 

\section{Related work}
Since the introduction of the Generative Pretrained Transformer (GPT), the field of large language models (LLMs) has witnessed a remarkable surge in innovation, fundamentally reshaping the landscape of natural language processing. This section offers an in-depth review of recent technical developments, spotlighting the key advancements that have influenced the direction of our research.

OpenAI's GPT series has been a trailblazer in the development of LLMs. GPT-1\cite{r2}  introduced the concept of unsupervised pre-training followed by fine-tuning, thereby setting a new benchmark for language understanding. GPT-2\cite{r3} further expanded on this approach by increasing the model's scale and training data, demonstrating remarkable zero-shot learning capabilities. GPT-3\cite{r4}, with its massive 175 billion parameters, revolutionized few-shot and one-shot learning. The subsequent releases of GPT-3.5 and GPT-4 have continued to push the boundaries, delivering even more impressive performance across a wide range of natural language tasks\cite{r9}.

Meta's LLaMA\cite{r10} has gained significant traction due to its focus on efficiency and scalability. It offers a more streamlined architecture compared to some of its predecessors while maintaining competitive performance. LLaMA has been widely adopted and adapted for various applications, highlighting its versatility and potential in different research and practical scenarios.
ByteDance's Doubao\cite{r22} is another powerful model. It is trained on extensive datasets, enabling it to understand and generate text across a diverse range of topics. Doubao excels in providing accurate and contextually relevant responses, making it a valuable tool for users seeking information and engaging in conversations. Alibaba's Qwen\cite{r11} is known for its high-performance natural language processing capabilities. It supports wide array of tasks, including text generation, question answering, and translation. Qwen is trained on a vast corpus of multi-modal data, which enhances its ability to handle complex language scenarios and provide comprehensive answers. DeepSeek\cite{r12} has also made a mark with its high-quality language generation and understanding. The model is designed to handle large-scale datasets and complex language structures, offering precise and coherent responses. DeepSeek's architecture is optimized for efficient training and inference, making it suitable for a variety of applications.Google's PaLM\cite{r13} is trained on massive datasets to capture a broad spectrum of language knowledge and semantic understanding. It has demonstrated excellent performance in tasks such as language translation, text summarization, and dialogue systems.

The Transformer architecture, the foundation of many modern LLMs, has undergone continuous refinement. Researchers have explored ways to enhance the self-attention mechanism. For instance, some studies have introduced multi-scale attention\cite{r14}, allowing the model to capture dependencies at different granularities. This approach enables the model to better handle long-range and short-range semantic relationships in text.Hybrid architectures combining Transformers with other neural network components have also been proposed. The integration of convolutional layers with Transformers\cite{r15} can capture local linguistic patterns more effectively. Convolutional filters can extract local features such as n - grams, complementing the global attention mechanism of Transformers.

To train large language models more efficiently, advanced training strategies have been developed. Curriculum learning\cite{r16} has been applied to LLMs, where the model is first trained on simpler tasks or data and gradually exposed to more complex ones. This approach can accelerate convergence and improve the model's generalization ability.Another significant development is the use of efficient optimization algorithms. Adaptive learning rate schedulers, such as AdamW\cite{r17}, have been widely adopted. AdamW adjusts the learning rate during training, helping the model converge faster and avoid overfitting. Additionally, techniques like gradient accumulation\cite{r18} allow for larger effective batch sizes when memory is limited, improving the stability of training.

Tokenization methods have evolved to better represent text for LLMs. Byte - Pair Encoding (BPE)\cite{r19} is a popular approach that splits words into subword units, reducing the vocabulary size while retaining the ability to represent rare or out-of-vocabulary words. Some recent works have proposed enhanced tokenization techniques that consider semantic and syntactic information\cite{r20}, resulting in more meaningful token representations.In terms of embeddings, researchers have explored ways to incorporate contextual information. Contextualized word embeddings, such as ELMo\cite{r21}, provide representations that vary depending on the surrounding text. This allows the model to capture the polysemous nature of words more effectively, improving performance in tasks that require fine - grained semantic understanding.In summary, the technical landscape of large language models has been rapidly evolving. 

Our proposed framework builds on these advancements, introducing novel mechanisms to further enhance the performance and capabilities of LLMs.

\begin{figure}[htbp]
    \centering
    \includegraphics[width=\textwidth]{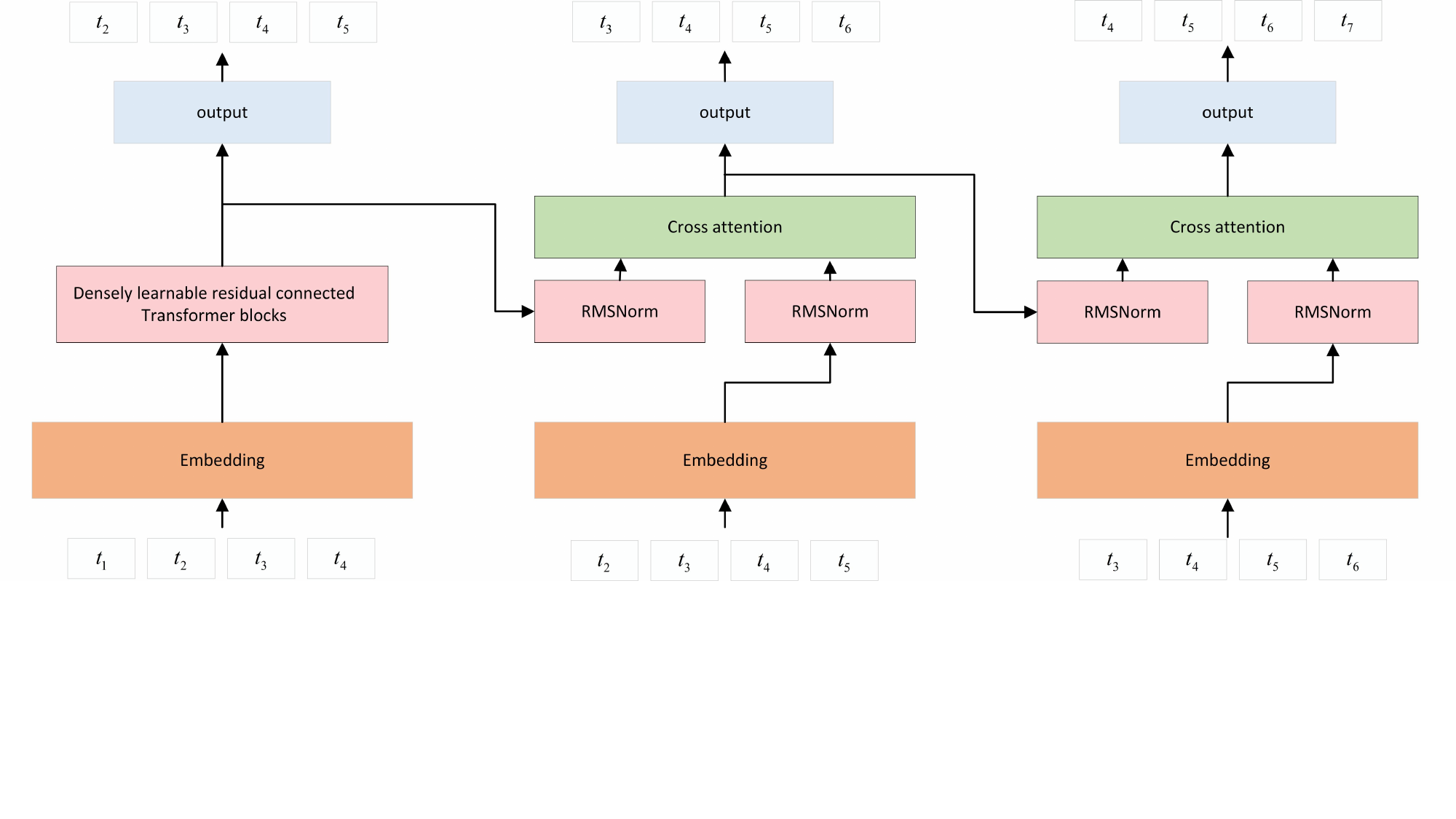}
    \caption{The architecture of KunlunBaize}
    \label{fig:your_label}
\end{figure}

\begin{figure}[htbp]
    \centering
    \includegraphics[width=\textwidth]{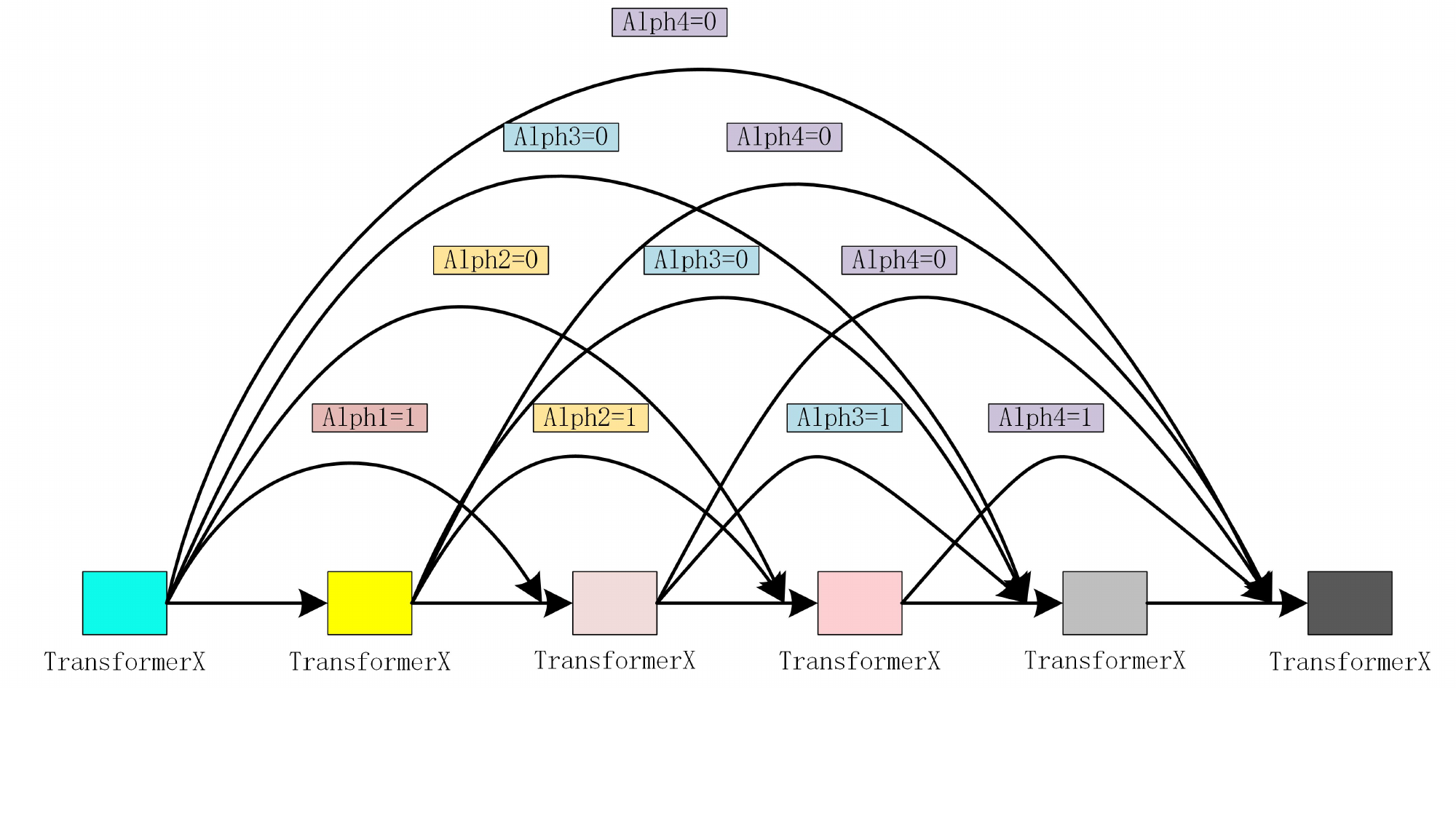}
    \caption{Densely leanable residual connected TransformerX blocks}
    \label{fig:your_label}
\end{figure}

\begin{figure}[htbp]
    \centering
    \includegraphics[width=\textwidth]{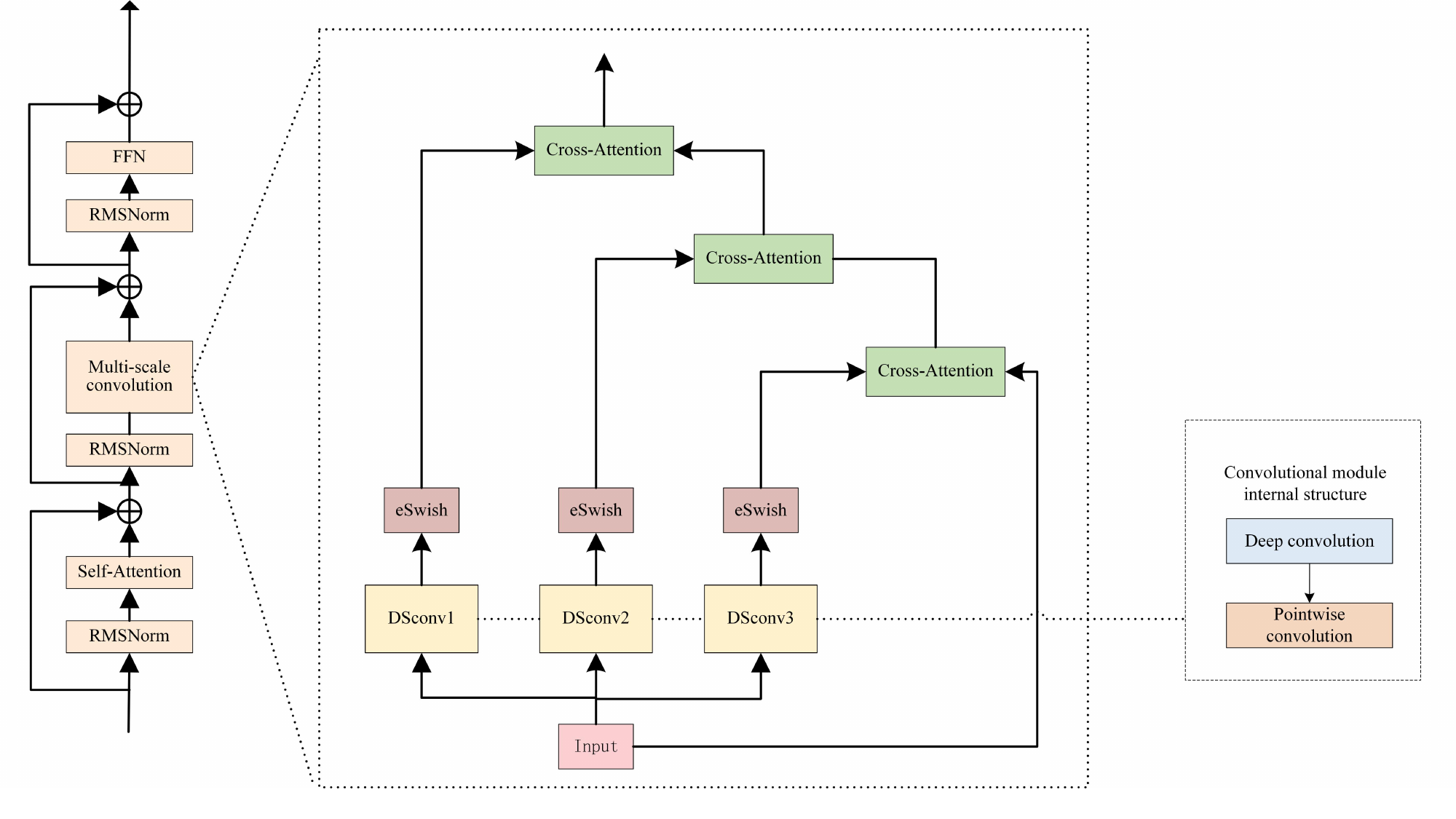}
    \caption{TransformerX}
    \label{fig:your_label}
\end{figure}

\section{Method}
In this section, we present our KunlunBaize Large Language Model, as shown in Figure 3. The entire framework mainly includes: (a) learnable dense residual skip connection mechanism for capturing the complex relationships between features at different levels; (b) TransformerX, a novel dual-scale convolutional transformer block capable of efficiently processing semantic information at different granularities, further enhanced by the adaptive activation function eSwish to improve its expressive power; and (c) multi-token prediction interaction module, which significantly improves data utilization efficiency and accelerates the model's speed during the inference phase. These modules work together to enhance the performance and efficiency of the large language model (LLM).

\subsection{Learnable Dense Residual Skip Connection Mechanism (LDRSCM)}

In the realm of deep - learning architectures, the ability to effectively capture hierarchical feature relationships and maintain stable information flow is crucial for model performance. We propose a novel \textbf{Learnable Dense Residual Skip-Connection Mechanism (LDRSCM)}, as depicted in Figure 4. This mechanism is designed to dynamically regulate the information - flow pathways within a neural network, endowing the model with enhanced capabilities to capture complex hierarchical feature relationships and facilitating more efficient information propagation across different layers.

\paragraph{Background and Motivation.}Traditional residual connections, first introduced in the ResNet architecture \cite{r44}, have been a cornerstone in deep neural network design. They have proven to be highly effective in alleviating the vanishing gradient problem, which is a major hurdle in training very deep neural networks. By providing a direct path for the gradient to flow back through the network, residual connections enable the training of architectures with a large number of layers. However, in the context of large-language models (LLMs), as the depth and complexity of the models increase, the limitations of fixed residual connections become more apparent.

In LLMs, different layers are responsible for capturing various levels of semantic and syntactic information. Fixed residual connections may not be able to optimally adapt to the changing importance of features at different layers. For example, in some language - related tasks, shallow-layer features such as basic word embeddings might be more relevant in certain contexts, while in other scenarios, deep-layer features that represent high-level semantic abstractions are crucial. Moreover, fixed residual connections often struggle to efficiently integrate shallow-layer information into deep-layer representations, which can lead to a loss of important low-level details in the final feature representations.

To overcome these limitations, we draw inspiration from DenseNet \cite{r45} and introduce learnable weights into the residual connection framework. The use of learnable weights allows the model to dynamically control the contribution of each layer's features to subsequent layers. This adaptability enables the model to better capture multi-scale dependencies in the data, as it can selectively emphasize or de-emphasize the influence of different layers based on the task at hand. Additionally, it improves feature reuse by allowing the model to re-utilize relevant features from earlier layers more effectively.

\paragraph{Mechanism Design.}Let $H_l$ denote the input to the $l$-th layer of the neural network. The output $H_{l + 1}$ of the $(l+1)$-th layer is computed using the following formula:

\begin{equation}
H_{l + 1}=\text{Layer}(H_l)+\sum_{i = 0}^{l}\alpha_i\cdot H_i
\end{equation}
where:

\begin{itemize}
    \item $\text{Layer}(\cdot)$ represents the transformation operation applied by the $l$-th layer. This can be a TransformerX layer, which is commonly used in modern language models for its ability to handle sequential data and capture long-range dependencies. In a TransformerX layer, operations such as multi-head self-attention, feed-forward networks, and normalization are typically involved to transform the input $H_l$ into a more meaningful representation.
    \item $H_i$ is the output of the $i$-th layer. These layer outputs represent the feature maps at different levels of the network, capturing various aspects of the input data. For example, in a language-model context, lower-layer outputs might represent word-level features, while higher-layer outputs could represent sentence-level or discourse-level features.
    \item $\alpha_i$ is a learnable weight parameter. This parameter dynamically adjusts the contribution of the $i$-th layer's features to the output of the $(l + 1)$-th layer. By learning these weights during the training process, the model can determine the optimal combination of features from different layers for the task at hand.
\end{itemize}

The learnable weights $\alpha_i$ are initialized 0 across all layers. This initial distribution provides a starting point for the model to learn the appropriate weights. During the training phase, these weights are optimized using the backpropagation algorithm. Backpropagation allows the model to calculate the gradients of the loss function with respect to the learnable weights and update them in a way that minimizes the loss. This iterative process of weight update enables the model to adapt to the specific characteristics of the training data and the task requirements.

\paragraph{Mathematical Formulation of Learnable Weights.}To ensure stable training and prevent any single skip connection from dominating the information flow, we apply a softmax normalization to the learnable weights $\alpha_i$. The softmax normalization is defined as follows:

\begin{equation}
\alpha_i=\frac{\exp(\alpha_i)}{\sum_{j = 0}^{l}\exp(\alpha_j)}
\end{equation}

This normalization operation has several important properties. Firstly, it ensures that the sum of all the weights $\sum_{i = 0}^{l}\alpha_i = 1$. This property provides a probabilistic interpretation of the contribution of each skip connection, where $\alpha_i$ can be seen as the probability of the $i$-th layer's features contributing to the output of the $(l + 1)$-th layer. Secondly, it helps in making the training process more stable by scaling the weights in a way that they are all comparable in magnitude. Without this normalization, some weights might grow much larger than others during training, leading to an unbalanced contribution of different skip connections and potentially causing the model to converge to a sub - optimal solution.

\paragraph{Gradient Analysis.}The gradient of the loss function $\mathcal{L}$ with respect to the learnable weights $\alpha_i$ is an important quantity in the training process, as it is used to update the weights during backpropagation. Using the chain rule of calculus, we can derive the following formula for the gradient:

\begin{equation}
\frac{\partial\mathcal{L}}{\partial\alpha_i}=\frac{\partial\mathcal{L}}{\partial H_{l+1}}\cdot\frac{\partial H_{l+1}}{\partial\alpha_i}=\frac{\partial\mathcal{L}}{\partial H_{l+1}}\cdot H_i
\end{equation}

The term $\frac{\partial\mathcal{L}}{\partial H_{l+1}}$ represents the gradient of the loss with respect to the output of the $(l + 1)$-th layer. This gradient indicates how changes in $H_{l+1}$ affect the loss. The term $\frac{\partial H_{l+1}}{\partial\alpha_i}$ is simply $H_i$, which means that the gradient of the loss with respect to $\alpha_i$ is proportional to the output of the $i$-th layer and the gradient of the loss with respect to $H_{l+1}$. This gradient is then used to update the learnable weights $\alpha_i$ in the direction that minimizes the loss, ensuring that the weights adapt to the specific requirements of the task at hand.

In Figure 6, we compare the accuracy of the fixed dense residual connection(FDRC), Standard Residual Connection(SRC) and our method on the AIME dataset.

\begin{figure}[h]
    \centering
    \includegraphics[width=\textwidth]{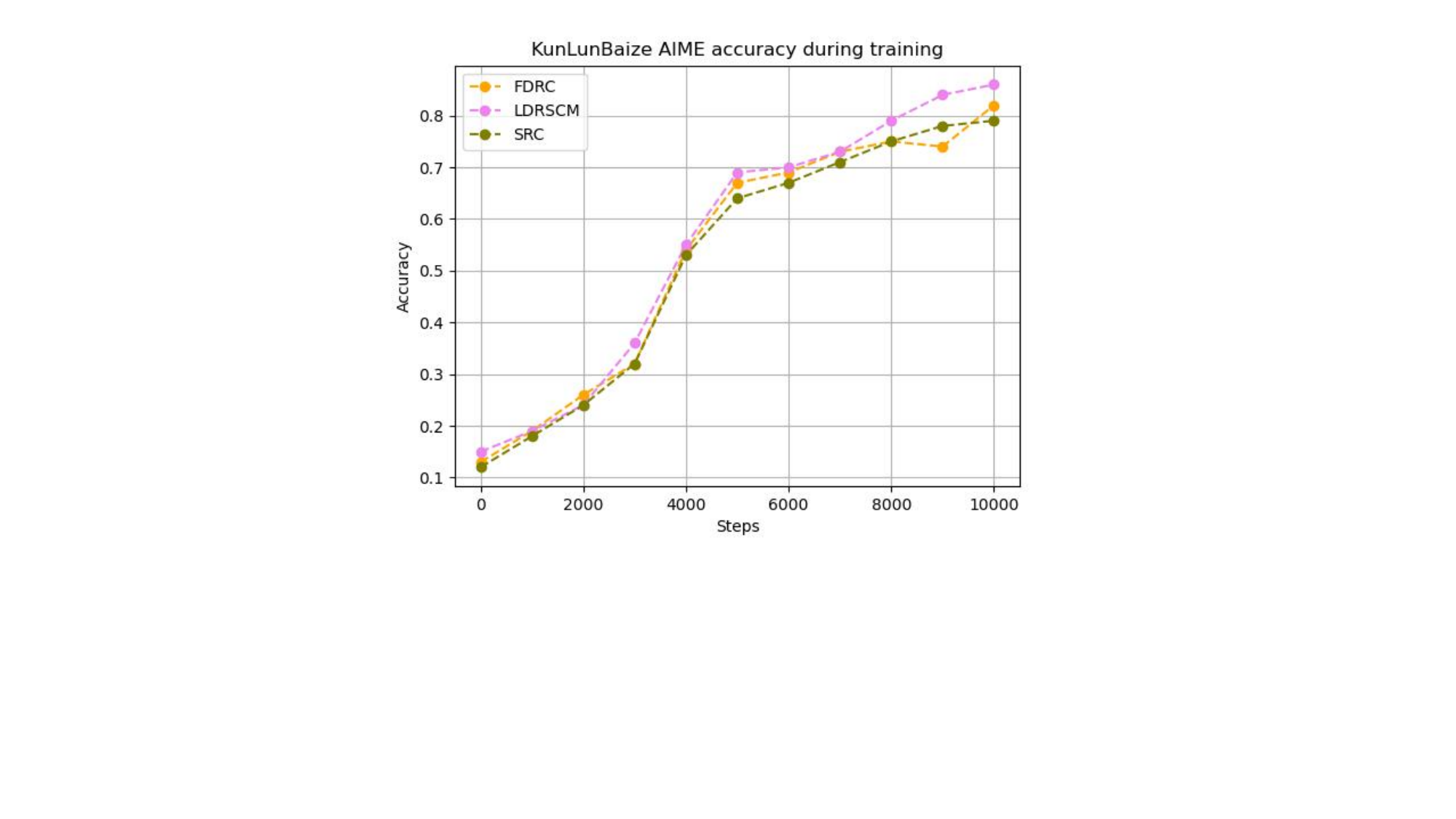}
    \caption{}
    \label{fig:your_label}
\end{figure}

\subsection{TransformerX: Multi-scale Convolution Transformer}
We introduce the TransformerX module, as shown in Figure 5. It augments information integration by incorporating additional convolution operations subsequent to the self-attention layer. The large-scale convolution aggregates semantic information from extended text segments, empowering the model to quickly grasp the macro-level themes and core logic. In contrast, the small-scale convolution focuses on the local word order and syntactic structures, capturing the subtle relationships and semantic collocations to precisely extract detailed information. Finally, we substitute the standard Swish activation function in the feedforward neural network with the adaptive eSwish function, which dynamically adjusts its parameters based on the semantic features of the input text. This adaptability enables the model to handle diverse contextual semantic expressions more effectively and enhances its capacity to model complex semantic relationships.

\subsubsection{Self adaptive Swish}
\begin{equation}
\text{Swish}(x)=x\cdot\sigma(x) \label{eq:swish}
\end{equation}
where $\sigma$ is the Sigmoid function. Although Swish performs well in multiple tasks, its fixed form may not fully meet the needs of different data distributions. Therefore, we propose an improved Swish activation function called eSwish, which is defined as:
\begin{equation}
\text{eSwish}(x)=x\cdot\sigma(\beta x) \label{eq:eswish}
\end{equation}
where $\beta$ is a learnable parameter used to dynamically adjust the shape of the activation function. By introducing $\beta$, the model can automatically optimize the nonlinear characteristics of the activation function according to the characteristics of the input data. We apply it to the FFN of the transformer block.

In Figures 7 and 8, we compared the changing trends of loss during the model training process between Swish and eSwish through experiments, as well as the accuracy on the AIME dataset, which proved the effectiveness of the eSwish we proposed.

\begin{figure}[h]
    \centering
    \includegraphics[width=\textwidth]{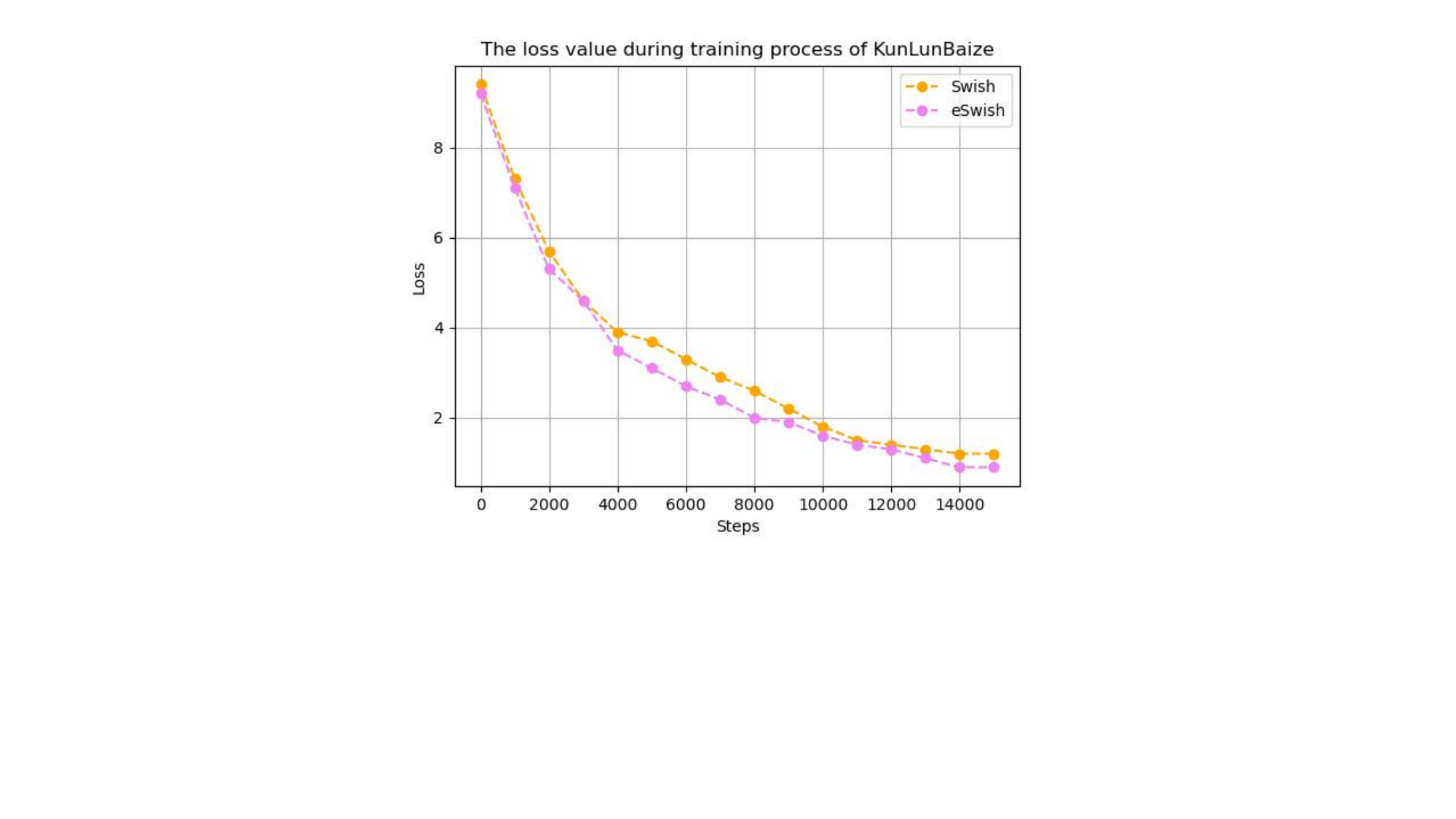}
    \caption{}
    \label{fig:your_label}
\end{figure}

\begin{figure}[h]
    \centering
    \includegraphics[width=\textwidth]{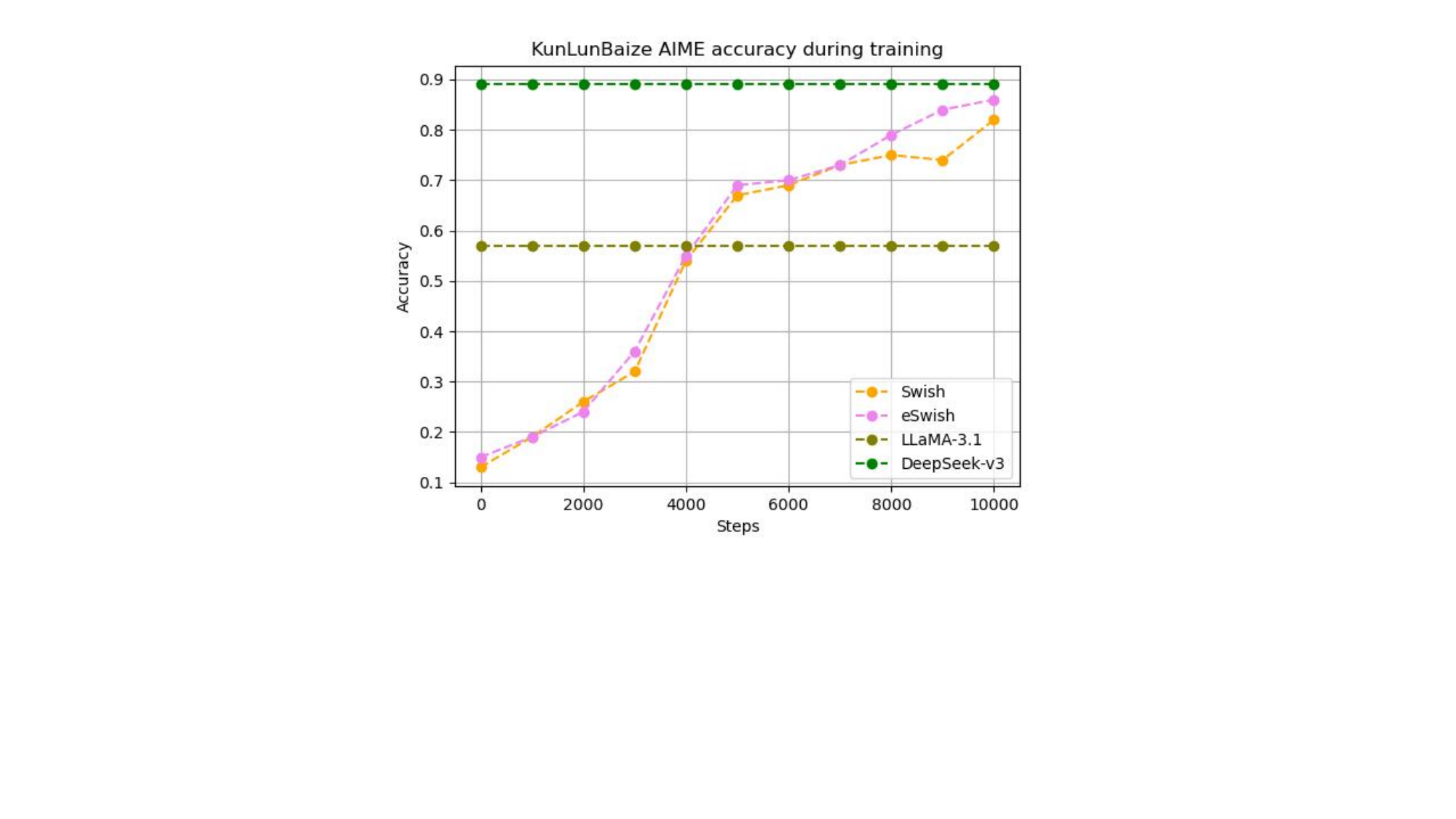}
    \caption{}
    \label{fig:your_label}
\end{figure}

\subsubsection{Multi-scale Convolution}

Let \(x\) be the input, and we make copies \(x_0, x_1, x_2, x_3, \dots\) of \(x\). 
For \(i = 1, 2, 3, \dots\), we define \(h_i =\text{eSwish} (\text{DSconv}_i(x_i))\). 
Among them, \(\text{DSconv}_i(x_i)\) represents Depthwise Separable Convolution, a convolutional operation employed in deep learning to reduce computational load and model parameters. It consists of two parts: Depthwise Convolution and Pointwise Convolution. We apply this two-dimensional convolutional operation to one-dimensional sequences. Specifically:

When dealing with one-dimensional sequences, assume the input sequence \(x_i\) has a shape of \([L, C]\), where \(L\) is the length of the sequence and \(C\) is the number of features or channels.

\paragraph{Depthwise Convolution for 1D Sequences.}For the depthwise convolution part, the depthwise convolution kernel \(K_{d}\) has a shape of \([k, C]\), where \(k\) is the length of the kernel. Instead of convolving over a two-dimensional spatial area as in the case of images, we convolve along the length of the one-dimensional sequence. The output \(y_{d}\) of the depthwise convolution is calculated as follows:

\begin{equation}
y_{d}[l, c] = \sum_{j=0}^{k-1} x_i[l + j, c] \cdot K_{d}[j, c]
\end{equation} 

for \(l = 0, \dots, L - k + 1\) and \(c = 1, \dots, C\). Here, for each channel \(c\) in the sequence, we slide the one-dimensional kernel \(K_{d}\) along the length of the sequence, multiplying and summing the corresponding elements.

\paragraph{Pointwise Convolution for 1D Sequences.}After obtaining the output \(y_{d}\) of the depthwise convolution, which has a shape of \([L - k + 1, C]\), we perform pointwise convolution. The pointwise convolution kernel \(K_{p}\) has a shape of \([1, C, N]\), where \(N\) is the number of output channels. The output \(y_{p}\) of the pointwise convolution is given by:

\begin{equation}
y_{p}[l, n] = \sum_{c=1}^{C} y_{d}[l, c] \cdot K_{p}[0, c, n]
\end{equation}

\begin{equation}
h=\text{eSwish}(y_p)
\end{equation}

for \(l = 0, \dots, L - k + 1\) and \(n = 1, \dots, N\). Let \(x_0^0 = x_0\), then

\begin{align}
x_0^1 &= \text{MLP}(\text{CrossAttention}(x_0^0, h_1))\\
x_0^2 &= \text{MLP}(\text{CrossAttention}(x_0^1, h_2))\\
&\cdots\\
x_0^{n} &= \text{MLP}(\text{CrossAttention}(x_0^{n - 1}, h_n))
\end{align}

Finally, we denote the output as \(x = x_0^{n}\)

\subsection{Multi-Token Prediction Interaction Module (MTPIM)}
Inspired by DeepSeek-V3\cite{r46}, we have adopted the multi-token prediction target method, as depicted in Figure 1. This approach extends the prediction scope of each position to multiple future tokens. By doing so, the training signals become denser, significantly enhancing the data utilization efficiency. We follow the method of predicting additional tokens within the sequence. On this basis, we have proposed a multi-token prediction interaction module. This module not only promotes the interaction between deep-level and shallow-level features but also ensures the integrity of the causal chain at each prediction depth. 

\subsubsection{Input Representation}
Let the input sequence be $T = [t_k, t_{k + 1},\cdots,t_{k + L}]$, where $k$ is the starting index of the current output sequence, and $L$ is the length of the sequence.

\subsubsection{Embedding Layer}
The input sequence $T$ is converted into a vector representation through the shared Embedding layer $E(\cdot)$:
\begin{equation}
h_{[k,k + L]} = E(T_{[k,k + L]}) \label{eq:embedding}
\end{equation}
where $h_{[k,k + L]}$ is the Embedding representation of the input sequence, and $E(\cdot)$ is the shared Embedding layer.

\subsubsection{TransformerX Encoding}
The Embedding representation $h_{[k,k + L]}$ is input into the TransformerX module for encoding to obtain the final feature representation:

\begin{equation}
h_{[k,k + L]}=\text{Blocks}(h_{[k,k + L]}) \label{eq:TransformerX_encoding}
\end{equation} 
where $\text{Blocks}(\cdot)$ representing the stacking of multiple TransformerX layers with dense residual connections.

\subsubsection{Multi-Token Prediction Interaction Module}
The multi-token prediction interaction module consists of $n$ interaction modules, and each module includes: a shared Embedding layer $E(\cdot)$, an independent cross-attention layer $C_n(\cdot)$ for each layer, and an independent output layer $O_n(\cdot)$ for each layer.

Shift the input sequence $T$ by one position to obtain $T_{[k + 1,k + L + 1]}$, and convert it into a vector representation through the Embedding layer $E(\cdot)$:
\begin{equation}
 h_{[k + 1,k + L + 1]} = E(T_{[k + 1,k + L + 1]}) \label{eq:input_shift}
\end{equation}
    
Perform RMSNorm normalization on $h_{[k,k + L]}$ and $h_{[k + 1,k + L + 1]}$ respectively:
\begin{align}
h_{[k,k + L]}&=\text{RMSNorm}(h_{[k,k + L]}) \label{eq:normalization1}\\
h_{[k + 1,k + L + 1]}&=\text{RMSNorm}(h_{[k + 1,k + L + 1]}) \label{eq:normalization2}
\end{align}
    
Input the normalized $h_{[k,k + L]}$ and $h_{[k + 1,k + L + 1]}$ into the cross-attention layer $C_n(\cdot)$ to obtain a new feature representation:
\begin{equation}
h_{[k + 1,k + L + 2]} = C_n(h_{[k,k + L]}, h_{[k + 1,k + L + 1]}) \label{eq:cross_attention}
\end{equation}
The calculation of the cross-attention layer can be expressed as:
\begin{equation}
C(Q, K, V)=\text{Softmax}(\frac{QK^T}{\sqrt{d_k}})V \label{eq:cross_attention_calculation}
\end{equation}
where $Q = h_{[k,k + L]}$, $K = V = h_{[k + 1,k + L + 1]}$
    
Repeat the above steps to gradually predict subsequent tokens. For the $j$-th interaction module, its output is:
\begin{equation}
h_{[k + j,k + L + j]} = C_j(h_{[k + j - 1,k + L + j - 1]}, h_{[k + j,k + L + j]}) \label{eq:recursive_prediction}
\end{equation}
    
The final output of the $j$-th layer is obtained through the independent output layer $O_j(\cdot)$:
\begin{equation}
y_{[k + j,k + L + j]} = O_j(h_{[k + j,k + L + j]}) \label{eq:output_layer}
\end{equation}
where $O_j(\cdot)$ is usually a linear layer that maps the features to the vocabulary space.

\subsubsection{Joint Loss Function}
To train the multi-token prediction interaction module, we adopt a joint loss function and sum the prediction losses of each interaction module with weights. Specifically, the loss function can be expressed as:
\begin{equation}
L_{\mathit{mpt} }=\sum_{j = 1}^{n}\gamma _jL^j_{\mathit{mpt} } \label{eq:joint_loss}
\end{equation}
where $L_{mpt}^j$ is the prediction loss of the $j$-th interaction module, usually using the cross-entropy loss:
\begin{equation}
L_{mpt}^j = -\sum_{i}\sum_{v}\mathbf{1}_{y_{i,v}}\log(P_{j,v}) \label{eq:cross_entropy_loss}
\end{equation}
where $y_{i,v}$ is the true label of the $i$-th target token, $P_{j,v}$ is the probability distribution predicted by the $j$-th interaction module, $V$ is the vocabulary size, and $\gamma_j$ is the corresponding weight coefficient used to balance the importance of different token prediction tasks.

\section{Post-Training}
\subsection{Reward Model}
We adhere to the dual reward model strategy adopted by OpenAI\cite{r47}, utilizing both a rule-based reward model and a narrative-based reward model.

\textbf{Rule-Based Reward Model}. The rule-based reward model allocates rewards based on predefined rules and logical criteria. These rules are typically designed to align with specific objectives, constraints, or desired behaviors. For example, in mathematical problem-solving tasks, the rule-based model might reward the agent for correctly applying formulas, following solution steps, or arriving at precise answers. At the same time, it can penalize deviations from logical reasoning or violations of mathematical rules. The strength of this model lies in its transparency and interpretability, as it clearly defines which behaviors are worthy of reward and which should be avoided.

\textbf{Narrative-Based Reward Model}\cite{r25}. The narrative-based reward model evaluates the agent's actions within the broader context of a story or narrative structure. This model takes into account the consistency, creativity, and long-term impact of the agent's decisions, rewarding behaviors that drive engaging and meaningful progression. For instance, in storytelling or interactive environments, the narrative-based model might reward the agent for developing intriguing plot points, maintaining character consistency, or fostering emotional engagement.

\subsection{Supervised Fine-Tuning (SFT) }
Supervised Fine-Tuning (SFT) is a crucial stage in the development of the KunLunBaize model. After the initial pre-training, SFT adapts the model to specific tasks and domains by leveraging labeled data. This process fine-tunes the model's parameters to enhance its performance on downstream tasks, making it more practical and useful in real-world applications. During this process, we conduct data screening and generation through rejection sampling. The specific process is as follows: After obtaining the reward model, it is necessary to collect appropriate data for subsequent supervised fine-tuning (SFT). At this time, a large number of candidate data will be generated from the existing model checkpoints. These data may contain various different answers and reasoning paths. For the reasoning data, it will be evaluated according to the preset rules and standards. If the data does not meet the requirements, such as unclear reasoning process, wrong answer or chaotic language expression, it will be rejected. Only the reasoning data that meet certain quality standards will be retained for subsequent model fine-tuning. In the process of collecting non-reasoning data, such as writing, factual question answering, self-awareness and translation tasks, similar principles will be adopted. After obtaining data from relevant data sources such as KunLunBaize, it will be screened to remove the parts that do not meet the requirements, ensuring that the non-reasoning data finally used for training has high quality and relevance.

\subsection{Offline Reinforcement Learning}

In the offline RL phase, we employ Direct Preference Optimization (DPO)\cite{r26} to fine-tune the model using preference pairs derived from SFT-trained prompts\cite{r27}. The pre-trained model generates diverse responses with varying temperature settings, which are evaluated by rule-based and narrative-based reward models to assess logical accuracy and contextual coherence, respectively. Preference pairs are constructed from the highest and lowest-scoring responses, and the model is trained using the AdamW optimizer (learning rate $1 \times 10^{-5}$, batch size 256) to align its outputs with these preferences. Validation on a held-out set ensures generalization, with early stopping applied to prevent overfitting. After several epochs, the model demonstrates improved performance, generating more contextually appropriate and user-preferred responses in tasks like customer service and creative writing\cite{r28}.

\section{ Experiments}
\subsection{Dataset Construction}

Based on previous experience in constructing LLM datasets, increasing the proportion of mathematics and programming in the corpus has often been shown to enhance the overall performance of models. This is because texts related to mathematics and programming typically exhibit a high degree of logical structure, which helps models better understand and generate complex reasoning processes\cite{r29}. To ensure the diversity and quality of the corpus, we incorporated a wide range of academic papers, technical documentation, and open-source code repositories, as well as high-quality educational resources, online courses, and carefully curated forum discussions. Additionally, we placed special emphasis on balancing cross-linguistic and cross-domain content to ensure the model's capability to handle multilingual tasks and knowledge across various fields. During the data cleaning and preprocessing phase, we employed multiple technical approaches, including deduplication, filtering low-quality content, correcting formatting errors, and removing noisy data. Furthermore, by combining manual review with automated tools, we further improved the accuracy and reliability of the corpus. Ultimately, we constructed a high-quality and highly diverse corpus containing 12.3 trillion tokens. This corpus not only covers a broad range of disciplines but also significantly enriches content related to mathematics, programming, and science, laying a solid foundation for model training.

\subsection{Model Parameter Configuration}
We largely adhere to the hyperparameter settings of LLaMA-3.1 405B\cite{r30}. Specifically, the TransformerX is configured with 126 layers and a hidden dimension of 16384. The Multi-Token Prediction Interaction Module consists of 6 layers, and all learnable parameters are randomly initialized with a standard deviation of 0.006. For the attention mechanism, we set the number of attention heads to 128. Additionally, the vocabulary size is set to 128256. This configuration ensures a robust and scalable architecture, balancing model capacity with computational efficiency.

\subsection{Training Hyperparameter Settings}
We largely follow the training configuration of llama3.1. The AdamW optimizer is employed, with hyperparameters set to $\beta_1 = 0.9$, $\beta_2 = 0.95$, and weight decay to 0.1. During the pre-training phase, the maximum sequence length is set to 4K, and KunlunKunlunBaize is pre-trained on 12.3 trillion tokens. For learning rate scheduling, we initially increase the learning rate linearly from 0 to $2.2 \times 10^{-4}$ over the first 2,000 steps. This is followed by maintaining a constant learning rate of $2.2 \times 10^{-4}$ until the model has processed 10 trillion training tokens. Subsequently, the learning rate is gradually reduced to $2.2 \times 10^{-5}$ over the next 1.8 trillion tokens, following a cosine decay curve. In the final 500 billion tokens of training, the learning rate remains constant at $2.2 \times 10^{-5}$ for the first 333 billion tokens, and then switches to a constant rate of $7.3 \times 10^{-6}$ for the remaining 167 billion tokens. The gradient clipping norm is set to 1.0.A batch size scheduling strategy is adopted, where the batch size is incrementally increased from 3,072 to 15,360 during the first 469 billion tokens of training, and then held constant at 15,360 for the remainder of the training process. This approach ensures a balanced trade-off between computational efficiency and model performance throughout the training lifecycle.

\subsection{Evaluation}

\subsubsection{Evaluation Benchmarks}
In the evaluation of model performance, we conducted a comprehensive comparison between KunlunBaize and several top-notch models. The models involved in the comparison include DeepSeek-V3, Qwen-2.5, LLaMA-3.1, MiniMax-01, Gemini-1.5-Pro, Claude-3.5-Sonnet, Gemini-2.0-Flash, and GPT-4o. This evaluation covered benchmark tests in multiple fields such as English, Code and Math, and Chinese to comprehensively assess the capabilities of KunlunBaize.

\subsubsection{Detailed Evaluation Configurations for KunlunBaize}
In standard benchmark tests, for tests like MMLU\cite{r31}, DROP\cite{r32}, GPQA\cite{r33}, and SimpleQA\cite{r34}, KunlunBaize adopts the evaluation prompts sourced from the simple-evals framework. In the zero-shot setting of MMLU-Redux, KunlunBaize utilizes the Zero-Eval prompt format as proposed by Li (2023)\cite{r35}. For other datasets, KunlunBaize strictly adheres to the original evaluation protocols and the default prompts provided by the dataset creators. This ensures a standardized and fair evaluation process, facilitating accurate comparisons with other models.

In the field of code and math benchmark tests, KunlunBaize uses a comprehensive evaluation approach. In code evaluation, the HumanEval-Mul dataset, which includes 8 mainstream programming languages such as Python, Java, Cpp, etc., is applied. \cite{r36}. When evaluating LiveCodeBench\cite{r37},  the data collected from August 2024 to November 2024 is used, and both the Chain-of-Thought (CoT) and non-CoT methods are-employed\cite{r38}.This allows for an in-depth understanding of its code- handling capabilities from different perspectives. In the evaluation of the Codeforces dataset\cite{r39}, KunlunBaize's performance is measured by the percentage of competitors. For the SWE-Bench verified evaluation, the agentless framework described by Reem Aleithan et al. (2024)\cite{r40} is used to objectively assess its capabilities in software engineering tasks. In Aider-related benchmark tests, the "diff" format is used to compare the differences between its outputs and the reference content, enabling a detailed analysis of its performance\cite{r41}.

In mathematical assessments, AIME and CNMO 2024 are evaluated with a temperature of 0.7. The results are averaged over 16 runs to ensure stability and reliability. This approach helps in reducing the impact of randomness in the model's outputs. For MATH-500, greedy decoding is employed. This decoding strategy allows KunlunBaize to make locally optimal decisions during the generation process, maximizing the chances of producing accurate and relevant mathematical solutions\cite{r42}.

In math evaluations, for AIME and CNMO 2024 tests, a temperature of 0.7 is set, and the results are averaged over 16 runs to ensure stability and reliability. For the MATH-500 test, the greedy decoding strategy is adopted to help the model make locally optimal decisions during the generation process, improving the accuracy of mathematical solutions. To ensure fairness in all benchmark tests, like other models, KunlunBaize has a maximum output limit of 8192 tokens for each benchmark test. By strictly implementing these evaluation configurations, a comprehensive and accurate assessment of KunlunBaize's capabilities can be achieved\cite{r43}.

\subsubsection{Evaluation Results}

\paragraph{English Benchmarks.} As shown in Tables 1 and 2, in the MMLU test, KunlunBaize achieved an EM score of 87.5, demonstrating strong competitiveness among top models and proving its extensive knowledge. In the DROP test, KunlunBaize obtained an F1 score of 89.1, performing well in long-text processing and surpassing models like Gemini-1.5-Pro and LLaMA-3.1. In the SimpleQA test, although it lagged behind GPT-4o, KunlunBaize still provided reliable factual information with an accuracy rate of 20.3. In the IF-Eval (Prompt Strict) test, KunlunBaize scored 88.5, outperforming most of the compared models, indicating excellent performance under specific prompts.

\paragraph{Code and Math Benchmarks.} In code benchmark tests, KunlunBaize excelled in algorithmic tasks. In the HumanEval-Mul test, its Pass@1 rate reached 63.7, exceeding models such as GPT-4o and Gemini-1.5-Pro. In the engineering task test of SWE-Bench Verified, KunlunBaize's resolution rate was 49.6, outperforming models like Claude-3.5-Sonnet and GPT-4o.

In math benchmark tests, KunlunBaize also stood out. In the AIME 2024 test, its Pass@1 rate was 37.9, surpassing many models. In the MATH-500 test, its EM score was 88.3, and in the CNMO 2024 test, its Pass@1 rate was 38.8. These results demonstrated its strong mathematical reasoning ability.

\paragraph{Chinese Benchmarks.} KunlunBaize also performed well in Chinese benchmark tests, surpassing all models. In the C-SimpleQA test, with an accuracy rate of 65.3, it outperformed DeepSeek-V3. In the CLUEWSC and C-Eval tests, KunlunBaize's EM scores were 92.4 and 86.9 respectively, showing obvious competitive advantages and suitability for various Chinese language application scenarios.

Overall, KunlunBaize performed outstandingly in different benchmark tests and was able to compete effectively with leading models. Its excellent performance across multiple fields, especially in Chinese tests, has laid a solid foundation for future applications. With further optimization, its comprehensive capabilities are expected to be enhanced.
```

\begin{table}[h]
    \centering
    \begin{tabular}{|p{4cm}|p{2cm}|p{2cm}|p{2cm}|p{2cm}|p{2cm}|}
        \hline
        Benchmark (Metric) & DeepSeek-V3 & Qwen-2.5 & LLaMA-3.1 & MiniMax-01 & KunlunBaize \\
        \hline
        Architecture & MoE & Dense & Dense & MoE & Dense   \\
        \hline
        Activated Params & 37B & 72B & 405B & - & 450B   \\
        \hline
        Total Params & 671B & 72B & 405B & - & 450B   \\
        \hline
        \multicolumn{6}{|c|}{English} \\
        \hline
        MMLU(EM) & 88.5 & 85.3 & 88.3 & \textbf{88.6} & 87.5   \\
        \hline
        MMLU-Redux(EM) & \textbf{89.1} & 85.6 & 88.9 & 88.0 & 85.9   \\
        \hline
        MMLU-Pro(EM) & 75.9 & 71.6 & \textbf{78.0} & 73.3 & 63.6   \\
        \hline
        DROP-shot(F1) & \textbf{91.6} & 76.7 & 88.3 & 83.7 & 89.1   \\
        \hline
        IF-Eval(Prompt Strict) & 86.1 & 84.1 & 86.5 & 84.3 & \textbf{88.5}   \\
        \hline
        GPQA-Diamond(Pass@1) & 59.1 & 49.0 & \textbf{65.0} & 49.9 & 57.8   \\
        \hline
        SimpleQA(Correct) & 24.9 & 9.1 & 17.1 & \textbf{38.2} & 20.3   \\
        \hline
        FRAMES(Acc.) & 73.3 & 69.8 & 72.5 & 80.5 & \textbf{86.2}   \\
        \hline
        LongBench v2 (Acc.) & 48.7 & 39.4 & 41.0 & 48.1 & \textbf{51.3}   \\
        \hline
        \multicolumn{6}{|c|}{Code} \\
        \hline
        HumanEval-Mul (Pass@1) & \textbf{82.6} & 77.3 & 81.7 & 86.9 & 91.7   \\
        \hline
        LiveCodeBench(Pass@1-COT) & \textbf{40.5} & 31.1 & 36.3 & 33.4 & 39.6   \\
        \hline
        LiveCodeBench(Pass@1) & 37.6 & 28.7 & 32.8 & 34.2 & \textbf{40.2}   \\
        \hline
        Codeforces(Percentile) & \textbf{51.6} & 24.8 & 20.3 & 23.6 & 50.2  \\
        \hline
        SWE Verified(Resolved) & 42.0 & 23.8 & \textbf{50.8} & 38.8 & 49.6  \\
        \hline
        Aider-Edit(Acc.) & 79.7 & 65.4 & \textbf{84.2} & 72.9 & 77.2  \\
        \hline
        Aider-Polygolt(Acc.) & \textbf{49.6} & 7.6 & 45.3 & 16.0 & 9.6  \\
        \hline
        \multicolumn{6}{|c|}{Math} \\
        \hline
        AIME2024(Pass@1) & \textbf{39.2} & 23.3 & 16.0 & 9.3 & 37.9  \\
        \hline
        MATH-500(EM) & \textbf{90.2} & 80.0 & 78.3 & 77.4 & 88.3  \\
        \hline
        CNMO2024(Pass@1) & \textbf{43.2} & 15.9 & 13.1 & 10.8 & 38.8  \\
        \hline
        \multicolumn{6}{|c|}{Chinese} \\
        \hline
        CLUEWSC(EM) & 90.9 & 91.4 & 85.4 & 87.9 & \textbf{92.4}  \\
        \hline
        C-Eval(EM) & 86.5 & 86.1 & 76.7 & 76.0 & \textbf{86.9}  \\
        \hline
        C-SimpleQA (Correct) & 64.8 & 48.4 & 51.3 & 59.3 & \textbf{65.3}  \\
        \hline
    \end{tabular}
    \caption{Benchmark Comparison of Different Models}
\end{table}

\begin{table}[h]
    \centering
    \begin{tabular}{|p{4cm}|p{2cm}|p{2cm}|p{2cm}|p{2cm}|p{2cm}|}
        \hline
        Benchmark (Metric) & Gemini-1.5-Pro & Claude-3.5-Sonnet & GPT-4o &  Gemini-2.0-Flash & KunlunBaize  \\
        \hline
        Architecture & MoE & - & - & - & Dense   \\
        \hline
        \ Activated Params & - & - & - & - & 450B   \\
        \hline
        \ Total Params & - & - & - & - & 450B   \\
        \hline
        \multicolumn{6}{|c|}{English} \\
        \hline
        MMLU(EM) & 86.8 & \textbf{88.3} & 87.2 & 86.5 & 87.5    \\
        \hline
        MMLU-Redux(EM) & - & 88.9 & 88.0 & - & 85.9   \\
        \hline
        MMLU-Pro(EM) & 75.8 & \textbf{78.0} & 72.6 & 76.4 & 63.6   \\
        \hline
        DROP-shot(F1) & \textbf{89.2} & 88.3 & 83.7 & - & 89.1   \\
        \hline
        IF-Eval(Prompt Strict) & 86.5 & 86.5 & 84.3 & 88.4 & \textbf{88.5}   \\
        \hline
        GPQA-Diamond(Pass@1) & 65.0 & \textbf{65.0} & 49.9 & 62.1 & 57.8   \\
        \hline
        SimpleQA(Correct) & 23.4 & 28.4 & \textbf{38.2} & 26.6 & 20.3   \\
        \hline
        FRAMES(Acc.) & - & 72.5 & 80.5 & - & \textbf{86.2}   \\
        \hline
        LongBench v2 (Acc.) & - & 41.0 & 48.1 & - & \textbf{51.3}   \\
        \hline
        \multicolumn{6}{|c|}{Code} \\
        \hline
        HumanEval-Mul (Pass@1) & 86.8 & 81.7 & 80.5 & 89.6 & \textbf{91.7}   \\
        \hline
        LiveCodeBench(Pass@1-COT) & - & 36.3 & 33.4 & - & \textbf{39.6}   \\
        \hline
        LiveCodeBench(Pass@1) & - & 32.8 & 34.2 & - & \textbf{40.2}   \\
        \hline
        Codeforces(Percentile) & - & 20.3 & 23.6 & - & \textbf{50.2}  \\
        \hline
        SWE Verified(Resolved) & - & 20.8 & 28.8 & - & \textbf{49.6}  \\
        \hline
        Aider-Edit(Acc.) & 63.3 & \textbf{84.2} & 72.9 & - & 77.2  \\
        \hline
        Aider-Polygolt(Acc.) & - & 5.3 & 6.0 & - & \textbf{9.6}  \\
        \hline
        \multicolumn{6}{|c|}{Math} \\
        \hline
        AIME2024(Pass@1) & - & 16.0 & 9.3 & - & \textbf{37.9}  \\
        \hline
        MATH-500(EM) & 74.4 & 78.3 & 74.6 & 83.9 & \textbf{88.3}  \\
        \hline
        CNMO2024(Pass@1) & - & 13.1 & 10.8 & - & \textbf{38.8}  \\
        \hline
        \multicolumn{6}{|c|}{Chinese} \\
        \hline
        CLUEWSC(EM) & - & 85.4 & 87.9 & - & \textbf{92.4}  \\
        \hline
        C-Eval(EM) & - & 76.7 & 76.0 & - & \textbf{86.9}  \\
        \hline
        C-SimpleQA (Correct) & 64.8 & 51.3 & 59.3 & 63.3 & \textbf{65.3}  \\
        \hline
    \end{tabular}
    \caption{Benchmark Comparison of Different Models}
\end{table}

\begin{table}[htbp]

\begin{center}
\begin{tabular}{c|cccccc}
\toprule  
\\
Model & 4k & 8k & 16k & 32k & 64k & 128k  \\
\midrule 

GPT-4o & 0.970  & 0.921  & 0.890 & 0.888 & 0.884 & 0.881\\
Claude-3.5-Sonnet & 0.965  & 0.960  & 0.957 & 0.950 & 0.952 & 0.938\\
Gemini-1.5-Pro & 0.962  & 0.960  & 0.960 & 0.958 & 0.938 & 0.917\\
MiniMax-Text-01 & 0.963  & 0.961  & 0.953 & 0.954 & 0.943 & 0.947 \\
KunLunBaize & 0.973  & 0.968  & 0.969 & 0.962 & 0.957 & 0.948 \\
\bottomrule 
\end{tabular}
\end{center}
\caption{We compared the capabilities of different models under varying context lengths}
\end{table}

As shown in Table 3, our model has achieved comparable or even better results than the current state of the art models at different context lengths. Notably, the baseline results of our model are mainly sourced from the works cited in \cite{r46,r48}.

\subsection{Ablation Experiments}
To further validate the effectiveness of each component in the proposed framework, we conducted a series of ablation experiments. The ablation experiments aimed to evaluate the impact of removing specific modules on the performance of the KunlunBaize model, providing insights into the contribution of each component to the overall model performance.

\subsubsection{Experimental Setup}
We designed three ablation scenarios by removing one key component at a time from the complete KunlunBaize model:

\paragraph{Removing the Learnable Dense Residual Skip Connection Mechanism.}We replaced the learnable dense residual skip connections with simple identity mappings, which meant that the output of each layer only depended on the current layer’s computation, ignoring the information from previous layers.
\paragraph{Removing the TransformerX Module.}In this setting, the TransformerX module was substituted with a standard Transformer layer without multi-scale convolutions and the adaptive eSwish activation function. The standard Transformer layer utilized the original Swish activation function and lacked the additional convolutional operations for aggregating semantic information at different scales.
\paragraph{Removing the Multi-Token Prediction Interaction Module.}We removed the multi-token prediction interaction module and reverted to the traditional single-token prediction method. This change meant that the model only predicted one future token at a time, rather than multiple tokens, thus reducing the data utilization and feature interaction during the inference process.

For each ablation experiment, we trained the modified models using the same dataset, hyperparameter settings, and training procedures as the original KunlunBaize model. After training, we evaluated the performance of the ablated models on the same set of benchmarks used for evaluating the complete KunlunBaize model, including English, Code and Math, and Chinese benchmarks.

\subsubsection{Results and Analysis}

\paragraph{Impact of Removing the Learnable Dense Residual Skip Connection Mechanism.}The model without the learnable dense residual skip connection mechanism exhibited a significant performance drop across all benchmarks. On the English benchmarks, the scores on MMLU, MMLU-Pro, DROP, and SimpleQA decreased notably. For instance, the F1 score on DROP dropped by 4.2 percentage points compared to the original KunlunBaize model. This decline indicated that without the ability to dynamically adjust the information flow and capture complex inter - level feature relationships, the model struggled to maintain its performance, especially in tasks requiring the understanding of long texts and complex semantic information. The vanishing gradient problem became more severe, affecting the training stability and the model’s ability to learn effectively.
\paragraph{Impact of Removing the TransformerX Module.}When the TransformerX module was removed, the model’s performance in handling semantic information at different granularities was severely affected. In the code and math benchmarks, the Pass@1 rate on HumanEval-Mul decreased by 1.6 percentage points, and the model’s performance on engineering tasks in SWE-Bench Verified also deteriorated. The absence of multi-scale convolutions made it difficult for the model to quickly grasp macro-semantics and extract detailed local information simultaneously. The fixed Swish activation function in the standard Transformer layer was less effective in adapting to diverse semantic expressions, resulting in a weakened ability to model complex semantic relationships.
\paragraph{Impact of Removing the Multi-Token Prediction Interaction Module.}The model without the multi-token prediction interaction module showed reduced data utilization efficiency and slower inference speed. On all benchmarks, the performance was inferior to that of the original KunlunBaize model. In the mathematical assessments, the average scores on AIME and CNMO 2024 decreased, indicating that the model’s ability to make accurate predictions and maintain a complete causal chain during inference was compromised. The reduced feature interaction between different layers led to a less comprehensive understanding of the input sequence, affecting the model’s performance in generating high-quality outputs.

\begin{table}[htbp]

\begin{center}
\begin{tabular}{ccc|ccccc}
\toprule  

$\mathbf{LDRSCM}$ & $\mathbf{MTPIM}$ & $\mathbf{TransformerX}$ & $MMLU$ & $MMLUPro$  &  $CSimpleQA$ & $GPQA$ &  \\
\midrule 

$\times$ & $\surd$  & $\surd$  & 86.6 & 59.8 & 65.1 & 53.4\\
$\surd$ & $\times$  & $\surd$  & 85.4 & 63.1 & 62.8 & 57.2\\
$\surd$ & $\surd$  & $\times$  & 85.6 & 62.6 & 64.2 & 55.1\\
$\surd$ & $\surd$  & $\surd$  & 87.5 & 63.6 & 65.3 & 57.8 \\
\bottomrule 
\end{tabular}

\end{center}

\begin{center}
\begin{tabular}{ccc|ccccc}
\toprule  

$\mathbf{LDRSCM}$ & $\mathbf{MTPIM}$ & $\mathbf{TransformerX}$ & $IFEval$ & $DROP$  &  $CNMO2024$ & $HumanEval$ &  \\
\midrule 

$\times$ & $\surd$  & $\surd$  & 85.6 & 84.9 & 36.9 & 61.4\\
$\surd$ & $\times$  & $\surd$  & 86.4 & 82.1 & 35.8 & 60.9\\
$\surd$ & $\surd$  & $\times$  & 83.6 & 83.6 & 38.2 & 62.1\\
$\surd$ & $\surd$  & $\surd$  & 88.5 & 89.1 & 38.8 & 63.7 \\
\bottomrule 
\end{tabular}
\caption{We demonstrate the impact of removing each module.}
\end{center}
\end{table}

\section{ Conclusions}
We propose a novel framework that incorporates a learnable dense residual skip connection mechanism, a TransformerX module, and a multi-token prediction interaction module. The learnable dense residual skip connection mechanism enables the model to dynamically adjust the information flow, effectively capturing complex inter-level feature relationships. This not only enhances the model's expressiveness but also alleviates the problem of gradient vanishing, ensuring stable training and remarkable performance. The TransformerX module innovatively integrates multi-scale convolutions and the adaptive eSwish activation function into the Transformer architecture. Large-scale convolutions help the model quickly grasp macro-semantics, while small-scale convolutions focus on local word order and syntax. The adaptive eSwish function adjusts its parameters based on the input semantics, enhancing the model's ability to handle diverse semantic expressions and complex relationships. The multi-token prediction interaction module improves data utilization and promotes feature interactions by expanding the prediction range. As a result, it accelerates the inference process and enhances the overall performance of the model. 

Experimental results validate the effectiveness of this framework. In various natural language processing tasks, the model equipped with this framework outperforms traditional models, demonstrating excellent performance in both semantic tasks and computational efficiency. 

 \bibliographystyle{elsarticle-num} 
 %\bibliography{ref2}

% Generated by IEEEtran.bst, version: 1.14 (2015/08/26)

\end{document}